\documentclass[conference]{IEEEtran}
\IEEEoverridecommandlockouts
\usepackage{cite}
\usepackage{amsmath,amssymb,amsfonts}
\usepackage{algorithmic}
\usepackage{graphicx}
\usepackage{textcomp}
\usepackage{xcolor}
\usepackage{multirow}

\def\BibTeX{{\rm B\kern-.05em{\sc i\kern-.025em b}\kern-.08em
    T\kern-.1667em\lower.7ex\hbox{E}\kern-.125emX}}
\begin{document}

\title{
\vspace*{0.2in}
A Quantitative Comparison of Centralised and Distributed Reinforcement Learning-Based Control for Soft Robotic Arms
}
\author{
\IEEEauthorblockN{
Linxin Hou\IEEEauthorrefmark{1}\IEEEauthorrefmark{2},
Qirui Wu\IEEEauthorrefmark{1}\IEEEauthorrefmark{3},
Zhihang Qin\IEEEauthorrefmark{4},
Neil Banerjee\IEEEauthorrefmark{2},
Yongxin Guo\IEEEauthorrefmark{2},
Cecilia Laschi\IEEEauthorrefmark{4}}
\IEEEauthorblockA{\IEEEauthorrefmark{2}Department of Electrical and Computer Engineering, National University of Singapore}
\IEEEauthorblockA{\IEEEauthorrefmark{3}Department of Computer Science, National University of Singapore}
\IEEEauthorblockA{\IEEEauthorrefmark{4}Department of Mechanical Engineering, National University of Singapore}
\IEEEauthorblockA{\IEEEauthorrefmark{1}These authors contributed equally to this work.}
\thanks{Corresponding author: Linxin Hou (Email: hou.linxin@u.nus.edu)}
\vspace{-1em}
}
\maketitle

\begin{abstract}
This paper presents a quantitative comparison between centralised and distributed multi-agent reinforcement learning (MARL) architectures for controlling a soft robotic arm modelled as a Cosserat rod in simulation. Using PyElastica and the OpenAI Gym interface, we train both a global Proximal Policy Optimisation (PPO) controller and a Multi-Agent PPO (MAPPO) under identical budgets. The study systematically varies the number of controlled sections $n$ and evaluates the performance of the arm to reach a fixed target in three scenarios: default baseline condition, recovery from external disturbance, and adaptation to actuator failure. Quantitative metrics used for the evaluation are mean action magnitude, mean final distance, mean episode length, and success rate. The results show that there are no significant benefits of the distributed policy when the number of controlled sections $n\le4$. In very simple systems, when $n\le2$, the centralised policy outperforms the distributed one. When $n$ increases to $4< n\le 12$, the distributed policy achieves higher final success rates under the same training budget. In these systems, distributed policy promotes a stronger success rate, resilience, and robustness under local observability and converges faster given the same sample size. However, centralised policies take much less time to train the same size of samples. These findings highlight the trade-offs between centralised and distributed policy in reinforcement learning-based control for soft robotic systems and provide actionable design guidance for future sim-to-real transfer in soft rod-like manipulators.
\end{abstract}


\section{Introduction}
Soft robotic arms allow adaptable interactions with uncertain environments \cite{b1, b2, b3}. However, controlling soft robotic arms remains difficult due to their high compliance, infinite configuration spaces, and strongly nonlinear dynamics \cite{b4, b5}. Model-based controllers rely on abstractions that are often too coarse or too brittle. Reduced-order rod models help, but identification and fidelity gaps remain, especially during fast motion and contact-rich tasks \cite{b6, b7, b8}. As a result, we turn to learning-based control that optimises performance directly from interaction rather than exact models. For soft systems, Reinforcement Learning (RL) is particularly attractive as policies can be trained to balance accuracy, smoothness, and energy consumption directly from interaction data \cite{b9, b10, b11}. However, real life data are scarce, exploration is risky, and compliance slows dynamics and complicates sensing, making sim-to-real algorithm development difficult \cite{b12, b13}.

\begin{figure}[!t]
    \centering
    \includegraphics[width=\linewidth, keepaspectratio]{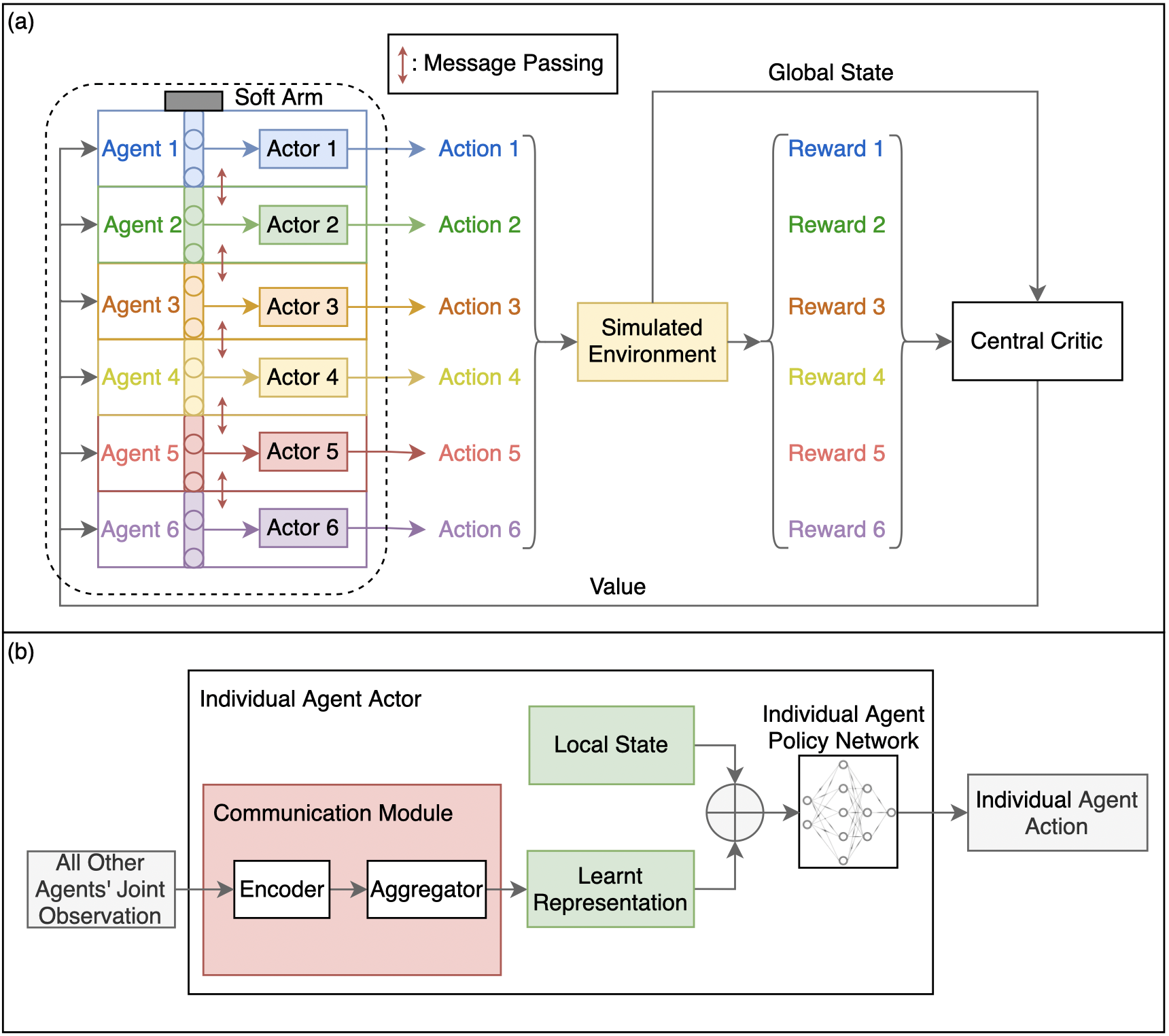}
    \caption{(a) n-agent distributed control architecture: the soft arm is partitioned into n agents; each actor issues an individual force command at its section. Message passing supports coordination. A central critic evaluates value from the global state (CTDE). (b) Individual agent block: a communication module forms a learned representation from other agents’ observations. It is fused with the local state and passed to the policy network to produce the agent's action.}
    \label{fig1} 
    \vspace{-2em} 
\end{figure}

In biology, distributed control is commonly seen in organisms under the phyla mollusca and cnidaria. For example, the octopus coordinates through distributed neural and muscular substrates which compute locally, adapt quickly, and avoid the burden of a single global planner \cite{b14, b15}. Local neural circuits in the arm compute and adapt with only partial, proximal information, while the central brain provides global modulation \cite{b14, b15}. This observation has informed a series of bio-inspired control designs for soft arms that prioritise local sensing–actuation loops and partial information, but not global state reconstruction \cite{b16, b17, b18, b19, b20, b21}. By using distributed systems, robustness and scalability may arise from the control architecture itself.

Inspired by the octopus, a soft arm can be partitioned into controllable sections that form a network of agents, each with local sensing–actuation loops and lightweight message passing. Multi-agent reinforcement learning (MARL) offers a natural formalism when a soft arm is discretised into sections, each acting as an agent. Message-encoded inter-agent communication mirrors neural signalling among arm ganglia, while the central critic abstracts the role of the integrative brain, enabling coordinated yet flexible behaviour. These biological parallels motivate our distributed formulation and inform the design of our communication and learning architecture.

Centralised designs exploit global observations and a joint policy. Distributed designs rely on independent learners with local observations and inter-agent communication \cite{b22, b23}. Each entails characteristic trade-offs. Centralised control typically coordinates well but scales poorly and can be brittle under partial observability while distributed control scales and tolerates local sensing, although it invites non-stationarity, coordination challenges, and higher training cost\cite{b24, b25, b26, b27, b28, b29}. 

Therefore, this paper targets design choice under controlled conditions. We present a quantitative evaluation of centralised and distributed MARL for a rod-like soft robotic arm performing a fixed-point reaching task in a simulated environment. Our results indicate that distributed control can produce resilience and robustness with faster convergence while centralised control achieves stronger coordination in simple systems. The crossover depends on the number of sections. Consequently, within the limits of our setting, we provide actionable guidance to select learning architectures in rod-like soft arm control.

\section{Methodology}
\subsection{System Description}
\begin{figure}[!t]
    \centering
    \includegraphics[width=\linewidth, keepaspectratio]{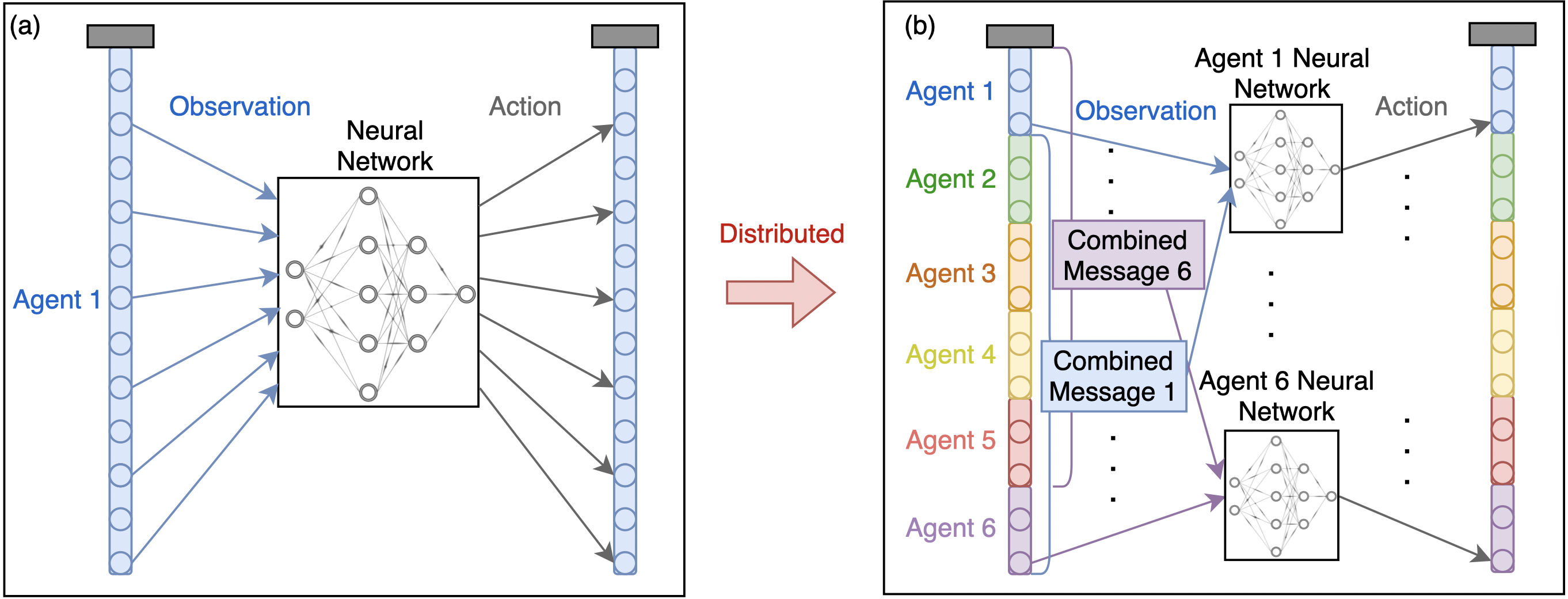}
    \caption{(a) Centralized policy: a single PPO controller receives global observations from the full rod and outputs joint actions to all forcing points. (b) Distributed policy: MAPPO with per-section agents. Each agent uses local observations (message aggregation) to produce its own action}
    \label{fig2} 
    \vspace{-1.5em} 
\end{figure}

We simulate a rod-like soft arm using PyElastica with the Cosserat-rod formulation \cite{b30}. A single elastic rod is discretised into 12 nodes and $n$ sections with $n$ forcing points where \( n \in \{2,3,4,6,8,12\} \). In the centralised setting, the whole rod is one agent that outputs a $n$-dimensional action to all forcing points. Proximal Policy Optimisation (PPO) is used in the centralised setting \cite{b31}. In the distributed setting, each section constitutes one agent (total $n$) trained with Multi-agent Proximal Policy Optimisation (MAPPO) under Centralised Training and Distributed Execution (CTDE) \cite{b32}\cite{b33}. Figure \ref{fig1} shows the distributed policy framework. Figure \ref{fig2} visualises the comparison of the two policies: left, a centralised PPO policy with global observations and a joint critic; right, $n$ individual actors with local observations and a centralised critic.

\subsection{Control Objective}
The task is to drive the tip of the rod $p^{\text{tip}}_t$ to the fixed target $g$ while encouraging a smooth, low-effort actuation. Figure \ref{fig3} visualises the evaluation framework used to assess the performance of the policy. Panel (a) shows the deformation and trajectory of the tip of the rod as it approaches the target over time, illustrating the typical behaviour of convergence under learnt control. Panel (b) outlines the three simulation scenarios designed to probe resilience and robustness, which will be detailed in Section III.B. Together, these visualisations clarify the progression of the reaching task and the conditions under which centralised and distributed controllers are compared.

\subsection{Distributed Policy}
The distributed multi-agent formulation adopts MAPPO under CTDE. The soft arm is partitioned into $n$ sections with $n$ forcing points, each controlled by one agent with an individual actor and a centralised critic (CTDE). The agent $i$ applies a bounded 3-D force $a_i$ in its section with individual and joint action spaces
\begin{equation}
\mathcal{A}_i=\{a_i\in\mathbb{R}^3 \mid -f_{\max}\le a_{i,j}\le f_{\max},\ j\in\{1,2,3\}\},\qquad
\end{equation}
\begin{equation}
\mathcal{A}=\prod_{i=1}^{n}\mathcal{A}_i,\ \mathbf{a}=(a_1,\dots,a_{n})\in\mathcal{A}.
\end{equation}
Each agent operates on a local observation vector that combines self-state, a shared task context, and compact summaries of the other agents:
\begin{equation}
o_i=[\text{id}_i,\ p_i,\ f_i,\ g,\ p_{-i},\ f_{-i}],
\end{equation}
where $\text{id}_i\!\in\!\mathbb{R}$ is a normalised identifier, $p_i\!\in\!\mathbb{R}^3$ the position of the agent node $i$, $f_i\!\in\!\mathbb{R}^3$ its applied force, $g\!\in\!\mathbb{R}^3$ the shared target and $p_{-i},f_{-i}$ are flattened stacks of the positions and forces of the other agents ($\mathbb{R}^{3(n-1)}$ each). 

Each agent encodes its local observation via an MLP encoder as \(m_i = f_{\text{enc}}(o_i;\theta_{\text{enc}})\), where \(f_{\text{enc}}:\mathbb{R}^{d_{\text{obs}}}\!\rightarrow\!\mathbb{R}^{d_{\text{msg}}}\).
Messages from other agents are mean-pooled and processed by an MLP aggregator \(c_i = f_{\text{agg}}(\bar{m}_i;\theta_{\text{agg}})\), where \(f_{\text{agg}}:\mathbb{R}^{d_{\text{msg}}}\!\rightarrow\!\mathbb{R}^{d_{\text{msg}}}\)
The policy conditions on the concatenated input, \[\pi_i(a_i \mid o_i, c_i) = \mathcal{N}\!\big(\mu_\pi([o_i; c_i]),\,\sigma_\pi\big),\] enabling distributed coordination. This yields a differentiable mean-field communication pipeline:
\begin{equation}
c_i
= f_{\text{agg}}\!\left(
\frac{1}{|N_{-i}|}\sum_{j\neq i} f_{\text{enc}}(o_j;\theta_{\text{enc}});
\,\theta_{\text{agg}}
\right).
\end{equation}

A global state $\mathcal{S}=\mathbb{R}^{6n+3}$,is available only to the centralised critic where $s=[p_1,\ldots,p_{n},\ f_1,\ldots,f_{n},\ g]=[\mathbf{p},\mathbf{f},g]$ and $p^{\text{tip}}_t$ is the position of the tip of the rod at time $t$. We define 
\begin{equation}
d^{\text{tip}}_t=\|p^{\text{tip}}_t-g\|_2,\qquad 
d^i_t=\|p^i_t-g\|_2,\qquad
\end{equation}

\begin{equation}
\Delta^{\text{global}}_t=d^{\text{tip}}_{t-1}-d^{\text{tip}}_t,\qquad \Delta^i_t=d^i_{t-1}-d^i_t
\end{equation}

and an agent-specific effort cost is thus
\begin{equation}
c^i_t=\frac{\|f^i_t\|_2}{f_{\max}}.
\end{equation}

To balance team progress with individual accountability, each agent receives a mixed global–local reward

\begin{equation}
r_t^{\text{global}}=\lambda_d\,\Delta^{\text{global}}_t,\qquad
r_t^{i,\text{local}}=\lambda_d\,\Delta^i_t,\qquad
\end{equation}
\begin{equation}
r_t^{i}=0.5\,r_t^{\text{global}}+0.5\,r_t^{i,\text{local}}-\lambda_a\,c^i_t,
\end{equation}
where $\lambda_d>0$ scales distance improvement (default $10.0$) and $\lambda_a>0$ penalises the magnitude of the action (default $10^{-3}$). This potential-based shaping preserves optimality while improving learning stability and credit assignment.

During training, each $\pi_{i,\theta_i}$ is updated by the MAPPO clipped policy gradient using advantages computed against $V_\phi(s)$. At test time, the critic is discarded and the agents act solely from $o_i$, giving scalable and fully distributed control. Each actor is updated with the PPO clipped surrogate objective and an entropy bonus, while the critic minimises the mean-squared temporal difference error calculated by the Generalised Advantage Estimation (GAE) \cite{b34}.


\subsection{Centralised Policy}
The centralised formulation treats the entire rod as a single agent that controls all $n$ forcing points jointly. The policy $\pi_{\theta}(\mathbf{u}\mid s)$ receives the global state $\mathcal{S}=\mathbb{R}^{6n+3}$, $s=[p_1,\ldots,p_{n},\; f_1,\ldots,f_{n},\; g]=[\mathbf{p},\mathbf{f},g]$,
and produces a joint action vector $\mathbf{u}=(a_1,\ldots,a_{n})\in\mathbb{R}^{3n}$ with element-wise bounds
\begin{equation}
\mathcal{A}=\Big\{\mathbf{u}\in\mathbb{R}^{3n}\ \big|\ -f_{\max}\le u_j \le f_{\max}\Big\}.
\end{equation}
At time step $t$, the team reward shared by all agents is
\begin{equation}
    r_t = \lambda_d \cdot \Delta_t - \lambda_a \cdot \bar{c}_t
\end{equation}
where $\Delta_t$ is the distance improvement, $d^{\text{tip}}_t$ is the Current tip-to-target distance, and $\bar{c}_t$ is the average control cost across all agents.

The actors $\pi_\theta^i(a_i\!\mid\!o_i)$ are parameter-shared between agents. Training uses PPO with a Tanh–Gaussian actor and a global value function $V_\phi(s)$. The centralised policy is trained under the same parameters as the distributed policy. The agent in the centralised policy follows the same parameters as one individual agent in the distributed policy.

\begin{figure}[!t]
    \centering
    \includegraphics[width=\linewidth, keepaspectratio]{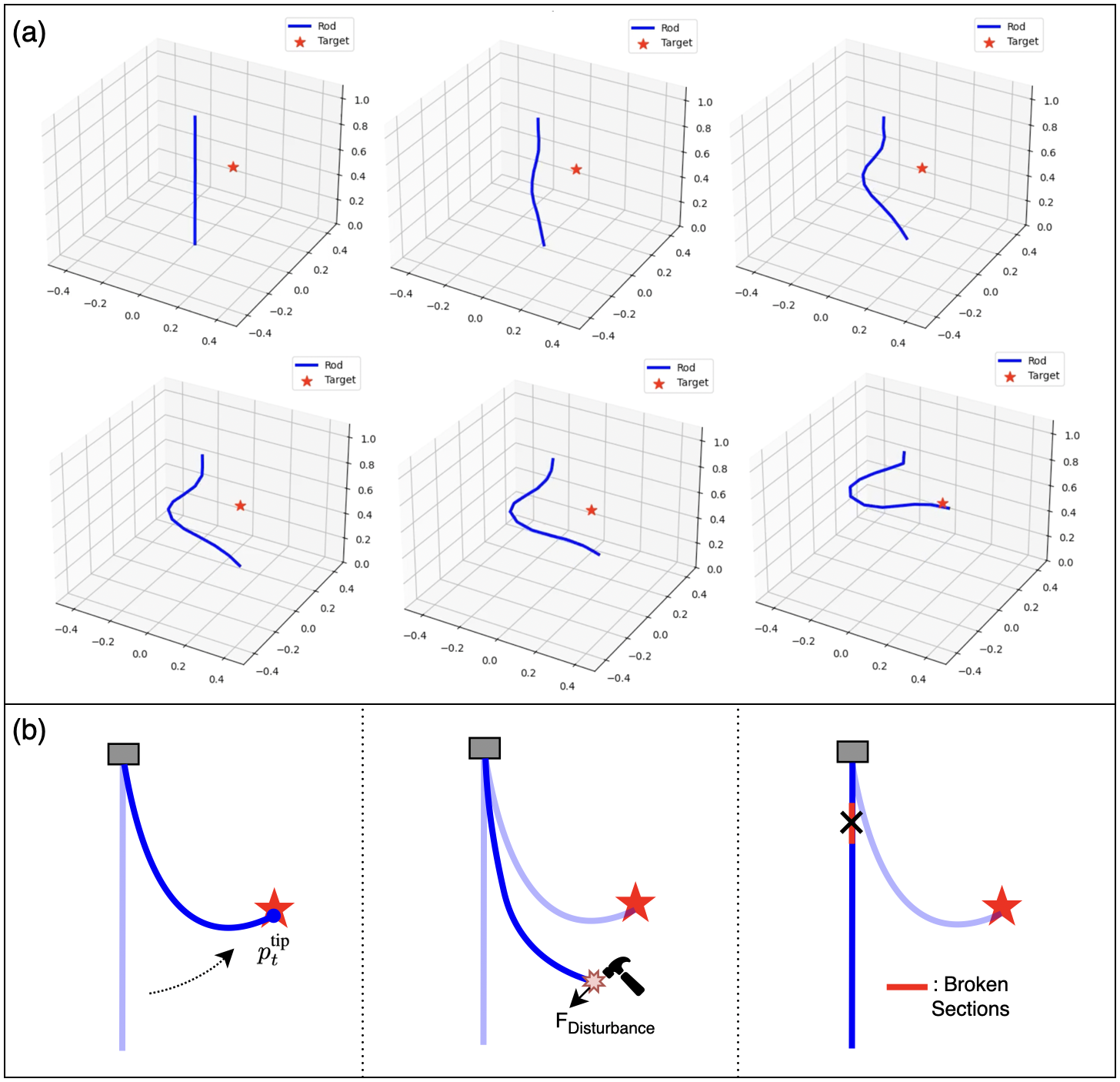}
    \caption{(a) Sequential snapshots of the simulated soft robotic arm (blue) in a fixed-point reaching task, where the rod tip $p^{\text{tip}}_t$ moves progressively toward the target $g$ (red star). (b) Illustration of the three testing scenarios.}
    \label{fig3} 
    \vspace{-2em} 
\end{figure}

\section{Simulation Setup}
All simulations were conducted in PyElastica, which implements the Cosserat-rod theory to simulate the continuous mechanics of the soft arm. The simulator was wrapped in an OpenAI Gym interface to enable on-policy rollouts and parallel evaluation under identical control and training conditions \cite{b35}. The soft arm modelled as a single elastic Cosserat rod is discretised into $n$ sections and actuated at $n$ forcing points in the soft arm. Episodes run in a 12-element rod environment with \texttt{control\_dt} \(=2\times 10^{-3}\,\mathrm{s}\), \texttt{max\_force} \(=15\,\mathrm{N}\), and \texttt{success\_radius} \(=0.03\,\mathrm{m}\).

\subsection{Policy Configurations}
Two architectures were evaluated under identical training budgets and hyperparameter settings: 
\begin{itemize}
    \item \emph{Centralized PPO (Single-Agent)}: A single actor–critic pair controls all actuation points jointly using global observations.
    \item \emph{Distributed MAPPO (Multi-Agent)}: Each section is an individual agent with a individual actor and a central critic under CTDE.
\end{itemize}

Both centralised PPO and distributed MAPPO architectures employ multilayer perceptrons with \texttt{Tanh} activations. The shared actor (centralised) and each individual actor (distributed) use two hidden layers of size \(128\), while the central critic in both architectures consists of two hidden layers of size \(256\). In the distributed MAPPO architecture, each agent additionally uses a two-layer message encoder (\(9 \rightarrow 64 \rightarrow 16\)) and aggregator (\(16 \rightarrow 32 \rightarrow 16\)) for inter-agent communication. Both architectures are trained separately for 10,000 updates using the same cluster of NVIDIA RTX A5000 GPUs.

Both the centralised policy (PPO) and the distributed policy (MAPPO) are trained for \texttt{total\_updates} \(=10{,}000\) using rollouts of \texttt{rollout\_steps} \(=2000\), and each update is optimised for \(5\) epochs with \texttt{minibatch\_size} \(=8192\). Each agent follows PPO with actor/critic learning rates \(3\times10^{-4}\) and \(10^{-3}\), clipping \(\epsilon=0.2\), discount \(\gamma=0.99\), GAE \(\lambda=0.95\), entropy coefficient \(0.01\), value loss coefficient \(0.5\), and gradient norm clipping at \(0.5\). Inter-agent communication uses fixed-width messages with dimension equal to \(16\), learnt end-to-end alongside the policy.

\subsection{Evaluation Scenarios}
To evaluate adaptability and robustness, trained controllers were tested in three conditions shown in Figure \ref{fig3}:
\begin{itemize}
  \item \emph{Scenario 1: Nominal Reaching}: Baseline condition with no external disturbance, used to assess convergence accuracy and smoothness.
  \item \emph{Scenario 2: Recovery from Disturbance}: At a fixed timestep, an impulse force $F$ is applied to the mid-span of the rod with fixed direction and magnitude. The controller must restore the tip to the target while minimising oscillations. Performance is measured by post-disturbance settling time.
  \item \emph{Scenario 3: Adaptation to Actuator Failure}: For $n=8$, one fixed actuator is disabled for the remainder of the episode. Successful adaptation is defined by compensatory coordination among the remaining agents and convergence to the goal.
\end{itemize}
The target point of the reaching task in all scenarios and the broken section in scenario 3 are unified for both architectures.

\subsection{Evaluation Metrics}
Both PPO and MAPPO were trained with identical horizon, discount factors, and optimisation schedules using Tanh–Gaussian policies and GAE. Each configuration was evaluated every 200 updates. This paper evaluated the trained models under 10,000 episodes in 3 scenarios. A mission ends when the tip-to-goal distance is satisfied \[\|p^{\text{tip}} - g\| \leq \varepsilon_{\text{goal}}\].The following metrics were recorded: mean action magnitude, mean final distance, mean episode length, and success rate.

\section{Result and Discussion}
\subsection{Training}
\begin{figure}[!t]
    \centering
    \includegraphics[width=\linewidth, keepaspectratio]{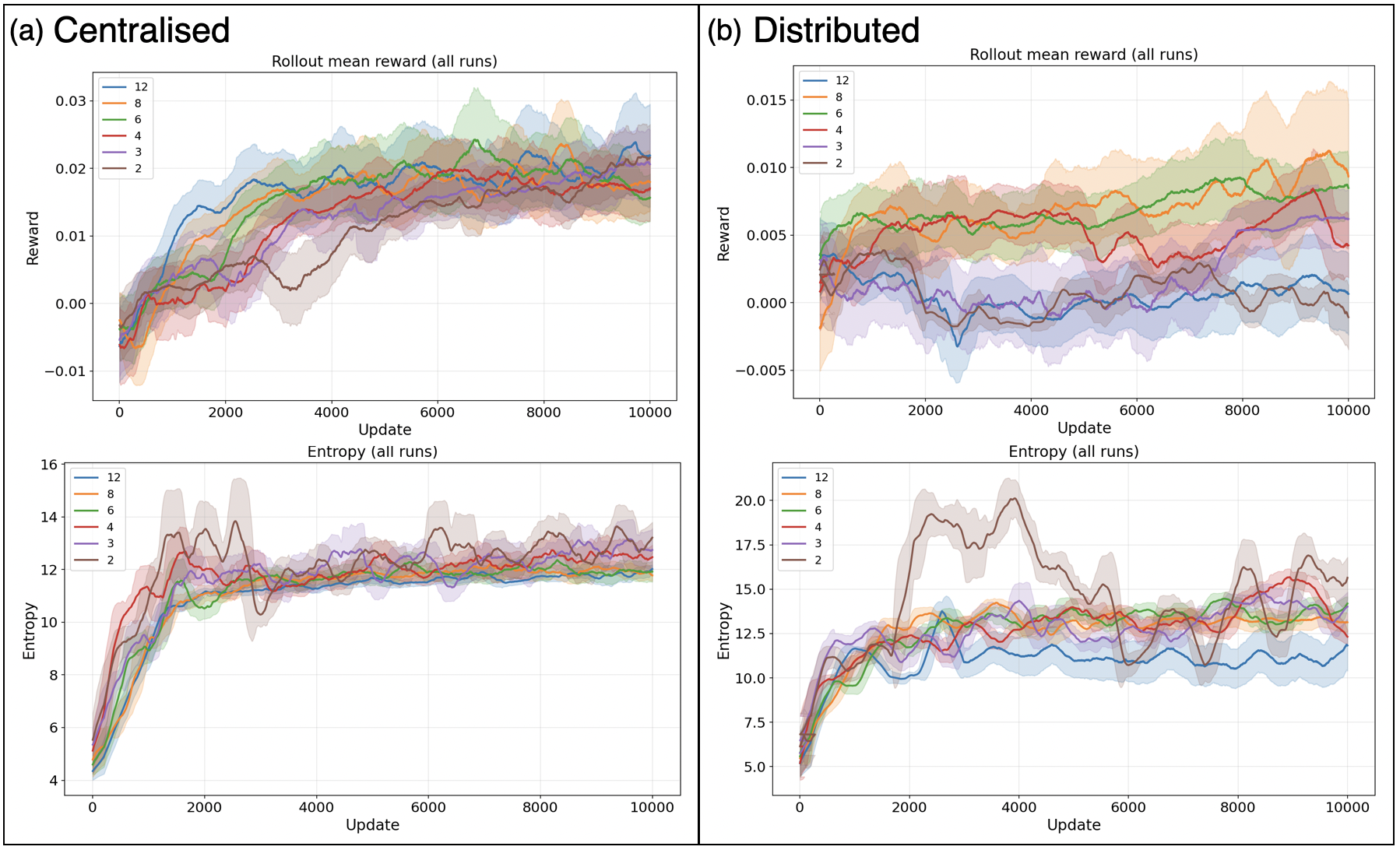}
    \caption{Training metrics of centralised (a) and distributed (b) policies over 10,000 updates for varying numbers of controlled sections $n$.}
    \label{fig4} 
    \vspace{-1.5em} 
\end{figure}

Figure \ref{fig4} summarises the training of 10,000 updates for varying $n$. The centralised policy converges faster at higher dimensions (from $n=2$ to $n=12$). This indicates effective global coordination. For simple systems where $n<4$, a distributed policy does not provide an advantage during training over a centralised control policy as it does not yield fast convergence. However, beyond this regime, when $4<n<12$, the distributed policy converges much faster, demonstrating better scalability with increasing system complexity. However, in a very complicated system ($n=12$ in our simulation), distributed policy struggles to converge. This may be caused by the message passing strategy used in this paper, causing inefficiency in communication in highly complex systems. When $2<n<8$, the two approaches demonstrate the same level of entropy after convergence, indicating a similar level of stability. When $n=2$, both policy networks show a clear increase in the entropy level after convergence, with the distributed one more evident. This reflects the control challenges of soft robotic control in highly under-actuated systems. The variance among rewards is also greater for distributed policy (especially $n=2, 3$), while centralised training exhibits tighter confidence bands at higher $n$, suggesting improved stability. Moreover, in early training, centralised runs often show a brief near-zero or mildly negative transient followed by smooth improvement, while distributed curves start closer to zero but display more pronounced mid-training oscillations around 3–7k updates, which is indicative of policy reorganisation.

\subsection{Evaluation}
PPO and MAPPO use different reward structures. Therefore, absolute reward values are not directly comparable between architectures. The paper uses reward curves only as within-method training diagnostics. Cross-method comparisons are based on task-level metrics evaluated under identical termination criteria.

Referring to Table \ref{tab:scenario1_results}, over 10,000 evaluation episodes, the distributed model achieves a higher success rate than the centralised model in the scenario 1 baseline as the complexity of the system increases (increasing \(n\)). For \(n \geq 6\), the distributed policy achieves near-perfect success with short episodes and small terminal errors. On the other hand, centralised policy fails in \(n=6\) to \(n=3\) with a success rate of \(0\%\) and mean episode length of \(=1000\). The centralised model is also less reliable and efficient in complex systems when \(n=8\) and \(n=12\), resulting in lower success rates with a longer mean episode length and a larger mean final distance. At the same time, when \(n \in \{3,4\}\), the centralised policy saturates with a success rate of 0 at the horizon, and the distributed model achieves modest success (\(33\%\) and \(10\%\)), indicating improved exploration but incomplete reliability. However, in a simple system the trend reverses. At \(n=2\), the centralised model is more reliable (\(73\%\)) than the distributed model (\(18\%\)). The mean magnitudes of the action increase with \(n\) for both methods, reflecting greater control authority in larger systems. In high \(n\), distributed models usually use slightly larger actions while reaching targets faster. Overall, these results indicate that distributed control scales more effectively for higher-section soft arms, while centralised control remains competitive only in very low complexity settings.

The zero-success entries for the centralised PPO at $n\in\{3,4,6\}$ should be understood in terms of convergence in the fixed training budget. At the checkpoint corresponding to the training time steps reported in Table~I/II, PPO in these intermediate settings $n$ had not yet stabilised. Its rollouts frequently terminate at the time horizon ($T_{\max}$) and the resulting policy behaviour remains in a transient, non-converged regime. In contrast, MAPPO reaches a clear performance plateau earlier under the same update/rollout budget, yielding both higher success rates and more stable behaviour at the same evaluation point (reflected by consistent episode lengths and terminal errors). Therefore, the observed $0\%$ success for PPO at these $n$ values reflects a slower convergence under the allocated budget rather than a fundamental limitation of centralised control. With additional training, PPO can improve, but this is beyond the scope of the present fixed-budget comparison.

\begin{table*}[ht]
\centering
\caption{Performance of centralized and distributed policies across different numbers of controlled sections ($n$) in \textbf{Scenario 1} (Baseline Nominal Reaching).}
\label{tab:scenario1_results}
\resizebox{\textwidth}{!}{
\begin{tabular}{|c|c|cccccc|}
\hline
\multirow{3}{*}{\centering\textbf{Metric}} & \multirow{3}{*}{\textbf{Policy}}
& \multicolumn{6}{c|}{\textbf{Scenario 1}} \\ \cline{3-8}
& &
\multicolumn{6}{c|}{\textbf{Number of Sections ($n$)}} \\ \cline{3-8}
& & \textbf{2} & \textbf{3} & \textbf{4} & \textbf{6} & \textbf{8} & \textbf{12} \\ \hline

\multirow{2}{*}{\centering\textbf{Mean Action Mag}} 
& Centralised & 31.38 & 39.22 & 45.60 & 52.64 & 41.02 & 63.16 \\ 
& Distributed & 30.32 & 39.84 & 47.55 & 58.30 & 66.84 & 64.11 \\ \hline

\multirow{2}{*}{\centering\textbf{Mean Final Distance}} 
& Centralised & $0.7423 \pm 0.2336$ & $0.2579 \pm 0.0519$ & $0.4417 \pm 0.0620$ & $0.0876 \pm 0.0203$ & $0.0494 \pm 0.0911$ & $0.1277 \pm 0.1424$ \\ 
& Distributed & $0.7886 \pm 0.3572$ & $0.4156 \pm 0.2859$ & $0.4545 \pm 0.2273$ & $0.0265 \pm 0.0002$ & $0.0286 \pm 0.0003$ & $0.0555 \pm 0.2254$ \\ \hline

\multirow{2}{*}{\centering\textbf{Mean Episode Length}} 
& Centralised & $742.3 \pm 233.6$ & $1000.0 \pm 0.0$ & $1000.0 \pm 0.0$ & $1000.0 \pm 0.0$ & $440.3 \pm 141.5$ & $676.9 \pm 261.0$ \\ 
& Distributed & $846.8 \pm 327.0$ & $727.9 \pm 387.8$ & $944.4 \pm 179.2$ & $101.0 \pm 0.0$ & $84.0 \pm 0.0$ & $80.7 \pm 131.3$ \\ \hline

\multirow{2}{*}{\centering\textbf{Success Rate (\%)}} 
& Centralised & 73.00 & 0.00 & 0.00 & 0.00 & 94.00 & 64.00 \\ 
& Distributed & 18.00 & 33.00 & 10.00 & 100.00 & 100.00 & 98.00 \\ \hline

\end{tabular}
}
\vspace{-1.5em}
\end{table*}

\begin{table*}[ht]
\centering
\caption{Performance of centralized and distributed policies across different numbers of controlled sections ($n$) in \textbf{Scenario 2} (Recovery from Disturbance) and \textbf{Scenario 3} (Adaptation to Actuator Failure).}
\label{tab:scenario2_3_results}
\resizebox{\textwidth}{!}{
\begin{tabular}{|c|c|cccccc|c|}
\hline
\multirow{3}{*}{\textbf{Metric}} & \multirow{3}{*}{\textbf{Policy}} 
& \multicolumn{6}{c|}{\textbf{Scenario 2}} & \multicolumn{1}{c|}{\textbf{Scenario 3}} \\ \cline{3-9}
& & \multicolumn{6}{c|}{\textbf{Number of Sections ($n$)}} & \multicolumn{1}{c|}{\textbf{Number of Sections ($n$)}} \\ \cline{3-9}
& & \textbf{2} & \textbf{3} & \textbf{4} & \textbf{6} & \textbf{8} & \textbf{12} & \textbf{8} \\ 
\hline

\multirow{2}{*}{\centering\textbf{Mean Action Mag}} 
& Centralised & 31.73 & 39.37 & 45.13 & 48.16 & 41.97 & 63.21 & 43.36 \\ 
& Distributed & 30.34 & 39.97 & 47.38 & 58.75 & 67.37 & 64.48 & 52.18 \\ \hline

\multirow{2}{*}{\centering\textbf{Mean Final Distance}} 
& Centralised & $0.0726 \pm 0.0768$ & $0.4405 \pm 0.0816$ & $0.4843 \pm 0.0376$ & $0.1068 \pm 0.0280$ & $0.0939 \pm 0.1721$ & $0.1598 \pm 0.1644$ &  $0.4125 \pm 0.1362$ \\ 
& Distributed & $0.8779 \pm 0.2518$ & $0.6637 \pm 0.1559$ & $0.3867 \pm 0.2004$ & $0.0294 \pm 0.0007$ & $0.0274 \pm 0.0003$ & $0.0693 \pm 0.2277$ & $0.0381 \pm 0.0376$ \\ \hline

\multirow{2}{*}{\centering\textbf{Mean Episode Length}} 
& Centralised & $758.7 \pm 237.7$ & $1000.0 \pm 0.0$ & $1000.0 \pm 0.0$ & $1000.0 \pm 0.0$ & $481.8 \pm 200.6$ & $745.7 \pm 254.0$ & $943.2 \pm 170.3$ \\ 
& Distributed & $932.5 \pm 228.9$ & $991.6 \pm 83.6$ & $963.2 \pm 115.0$ & $100.0 \pm 0.2$ & $87.0 \pm 0.0$ & $104.6 \pm 182.8$ & $382.6 \pm 182.4$ \\ \hline

\multirow{2}{*}{\centering\textbf{Success Rate (\%)}} 
& Centralised & 43.36 & 0.00 & 0.00 & 0.00 & 87.00 & 57.00 & 10.00 \\ 
& Distributed & 8.00 & 1.00 & 10.00 & 100.00 & 100.00 & 96.00 & 94.00 \\ \hline

\end{tabular}
}
\vspace{-2em}
\end{table*}

In scenario 2 of Table \ref{tab:scenario2_3_results}, under external disturbances, the distributed controller consistently exhibits better recovery compared to the centralised baseline, especially as the system scales. For \(n\geq 6\), the distributed policy achieves again near-perfect success with short episodes and small terminal errors (e.g., \(n=6\)). With lower section numbers, the advantage is mixed. The centralised model is more reliable at \(n=2\) (\(73\%\) vs. \(18\%\)), while for \(n\in\{3,4\}\) both are weak, but the distributed model remains slightly better (\(1\%\)–\(10\%\) vs. \(0\%\)). The mean magnitudes of action increase with \(n\) for both methods, with distributed employing larger commands at high \(n\) (e.g. \(67.4\) vs. \(42.0\) at \(n=8\)), consistent with aggressive, yet effective recovery. Overall, these results indicate that distributed policy is markedly more disturbance-resilient, producing faster, more accurate recoveries as system complexity grows.

Both the centralised and distributed models show good performance in scenario 1 when $n=8$. Therefore, these two models are selected to be tested for scenario 3 as shown in the last column of Table \ref{tab:scenario2_3_results}. With \(n=8\) in scenario 3, the distributed model shows strong fault tolerance, achieving a success rate \(94\%\) versus \(10\%\) for the centralised model. The distributed model completes episodes substantially faster (mean length \(382.6 \pm 182.4\) vs. \(943.2 \pm 170.3\)) and with a far smaller terminal error (\(0.0381 \pm 0.0376\) vs. \(0.4125 \pm 0.1362\)). The distributed policy also employs a higher mean action magnitude (52.18 vs. 43.36), suggesting more assertive compensation for the failed section. Overall, these results indicate that under actuator failure the distributed approach adapts effectively and shows strong robustness to maintain task performance, while the centralised model frequently stalls near the time horizon and fails to reach the target.

\section{Conclusion and Future Work}

This paper quantitatively compared centralised PPO and distributed MAPPO controllers for a Cosserat-rod soft arm across increasing numbers of controlled sections and three testing scenarios. The results indicate that there are no significant advantages of distributed policy in a simple system when the robot has few controlled sections $n$. As $n$ grows and the robot becomes more complex, the distributed approach achieves higher success rates, stronger resilience to disturbances, and robustness to actuator failures. The distributed approach also achieves faster convergence under the same sample size, while the centralised policy is quicker to train under the same sample size. Overall, these findings show clear trade-offs between centralised and distributed policies in RL-based control of compliant, distributed soft manipulators and offer pragmatic guidance for sim-to-real design of rod-like soft arms.

However, our study used a relatively small total number of training episodes, which constrains statistical power and may understate late-stage performance differences. Moreover, the current message-passing mechanism of the distributed policy appears suboptimal at high section counts, suggesting communication inefficiencies that can limit scalability. Despite these constraints, our results consistently support the above conclusions about the resilience, robustness, and convergence-speed benefits of distributed control in larger systems. Future work should focus on inter-agent communication to cut overhead and improve stability in high complexity robot. In addition, stronger distributed algorithms beyond MAPPO will be explored and designed to enhance coordination.

\section*{Acknowledgment}
This work was supported by Start-up grant RoboLife (Soft Robots with morphological adaptation and life-like abilities), DESTRO (Dextrous, strong yet soft robots), ITALY–SINGAPORE grant, MAE (Italy) and A*STAR (Singapore), Grant \#R22I0IR124, and Bridging fund (AI-Driven Soft Robots for Marine and Unstructured Environments).

\end{document}